\documentclass[review]{elsarticle}

\usepackage{hyperref} 

\journal{Journal of \LaTeX\ Templates}

\usepackage{bbm}
\usepackage{amsfonts}
\usepackage{graphicx}
\usepackage{amsmath}
\usepackage{amssymb}
\usepackage{booktabs}
\usepackage{bbding}
\usepackage{color}

\usepackage{bm}
\usepackage[lined,boxed,ruled]{algorithm2e}

\begin{document}

\begin{frontmatter}

\title{DSLA: Dynamic smooth label assignment for efficient anchor-free object detection
}

\author[iacas]{Hu Su\footnotemark[1]}
\author[iacas]{Yonghao He\footnotemark[1]}
\author[martU,normU]{Rui Jiang}
\author[iacas]{Jiabin Zhang}
\author[iacas]{Wei Zou\footnotemark[2]}
\author[techB]{Bin Fan\footnotemark[2]} 


\address[iacas]{Institute of Automation, Chinese Academy of Sciences, Beijing 100190, China.}
\address[martU]{College of Information Engineering, Shanghai Maritime University, Shanghai 201306, China.}
\address[normU]{School of Computer Science and Technology, East China Normal University, Shanghai 200062, China.}
\address[techB]{School of Automation and Electrical Engineering, University of Science and Technology Beijing, Beijing 100083, China.}

\begin{abstract}
Anchor-free detectors basically formulate object detection as dense classification and regression. For popular anchor-free detectors, it is common to introduce an individual prediction branch to estimate the quality of localization. The following inconsistencies are observed when we delve into the practices of classification and quality estimation. 
Firstly, for some adjacent samples which are assigned completely different labels, the trained model would produce similar classification scores. This violates the training objective and leads to performance degradation. Secondly, it is found that detected bounding boxes with higher confidences contrarily have smaller overlaps with the corresponding ground-truth.
\footnotetext[1]{These authors contributed equally.}
\footnotetext[2]{Corresponding authors.}
Accurately localized bounding boxes would be suppressed by less accurate ones
in the Non-Maximum Suppression (NMS) procedure. To address the inconsistency problems, the Dynamic Smooth Label Assignment (DSLA) method is proposed. Based on the concept of centerness originally developed in FCOS, a smooth assignment strategy is proposed. The label is smoothed to a continuous value in $[0, 1]$ to make a steady transition between positive and negative samples. Intersection-of-Union (IoU) is predicted dynamically during training and is coupled with the smoothed label. The dynamic smooth label is assigned to supervise the classification branch. Under such supervision, quality estimation branch is naturally merged into the classification branch, which simplifies the architecture of anchor-free detector. Comprehensive experiments are conducted on the MS COCO benchmark. It is demonstrated that, DSLA can significantly boost the detection accuracy by alleviating the above inconsistencies for anchor-free detectors. Our codes are released at https://github.com/YonghaoHe/DSLA.
\end{abstract}


\end{frontmatter}


\section{Introduction}
Convolutional neural networks (CNNs) have been widely adopted in computer vision tasks, including category classification \cite{ResNet:He}, object detection \cite{RCNN:Girshick, Point2Seq:Xue}, semantic segmentation \cite{FCN:Long} and other related tasks such as entity connection inference \cite{FGCN:Yang, DGCN:Yang} and cross-modality understanding \cite{Cross-modality:Huang}. Specially, object detection is a fundamental problem in computer vision, which aims to predict the locations of bounding boxes and corresponding category labels in an image. Since RCNN \cite{RCNN:Girshick}, deep-learning-based object detection has attracted much attention along with its wide applications in fields such as industrial detection \cite{QFasterRCNN:Zhang}, video analysis \cite{PSLA:Guo}, text recognition \cite{GASTD:Wang}, and aerial image \cite{RIM:Wang}. Existing deep-learning-based detectors could be roughly divided into anchor-free and anchor-based categories.
As popularized by Faster R-CNN \cite{FasterRCNN:Ren}, mainstream detectors such as SSD \cite{SSD:Liu}, RetinaNet \cite{Focalloss:Lin} and YOLO v2, v3 \cite{YOLO9000:Redmon} generally rely on a set of predefined anchor boxes to enumerate possible locations, scales and aspect ratios for the objects. Despite their promising performances, the detectors are limited to the design of anchor boxes. Recently, anchor-free detectors have gradually led the trend of object detection, which directly learn object possibility and the bounding box coordinates without anchor reference. Compared with anchor-based counterparts, anchor-free detectors get rid of the hyper-parameters and complicated computations related to anchor boxes, making the training process considerably simpler.
YOLOv1 \cite{YOLO:Redmon} is a popular anchor-free detector. Instead of using anchor boxes, YOLOv1 directly predicts bounding boxes at the points near the center of objects. CornerNet \cite{CornerNet:Law} and CenterNet \cite{CenterNet:Duan} adopt keypoint-based detection pipeline, which detects a pair of corners of a bounding box and groups them to form the final detected bonding box. FCOS \cite{FCOS:Tian}, CenterNet \cite{CenterNet:Zhou} and FoveaBox \cite{FoveaBox:Kong} formulate object detection in a per-pixel prediction fashion.

In order to train the detectors, defining positive and negative samples is a necessary yet important procedure which directly influences the training efficiency and thus the performance. The issue needs to be carefully considered especially for anchor-free detectors while anchor-based detectors divides the anchors into positive and negative samples according to Intersection-over-Union (IoU) values. Previous anchor-free detectors commonly adopt a single fixed division criterion. That is, positive and negative samples are divided according to hand-crafted rules and several predefined thresholds. For example, YOLOv1 \cite{YOLO:Redmon} divides the input image into a grid. If the center of an object falls into a grid cell, then that grid cell is regarded as positive and is responsible for detecting that object. CornerNet \cite{CornerNet:Law} only regards ground-truth (\textit{gt}) corner locations as positive and all other locations are negative. But the loss for negative locations within a radius of the positive locations is down-weighted. FCOS \cite{FCOS:Tian} and Foveabox \cite{FoveaBox:Kong} treat the locations within the center region or the bounding box of any \textit{gt} object as positive candidates. However, such static strategies could not adapt to various shapes and attitudes of objects to always provide the optimal positive/negative division. Dynamic assignment strategies are proposed. ATSS  \cite{ATSS:Zhang} proposes to set the division boundary for each \textit{gt} based on the statistics of IoU values. OTA \cite{OTA:Ge} attempts to find the globally best division strategy by solving optimal transport problem. Unfortunately, most of the methods are anchor-based and cannot be directly applied to anchor-free detectors. Meanwhile, the inconsistency occurring in anchor-free detectors has not been paid sufficient attention, which will be detailed in the following with FCOS as the example.

\begin{figure}[htbp]
\centering
\includegraphics[width=4.5in]{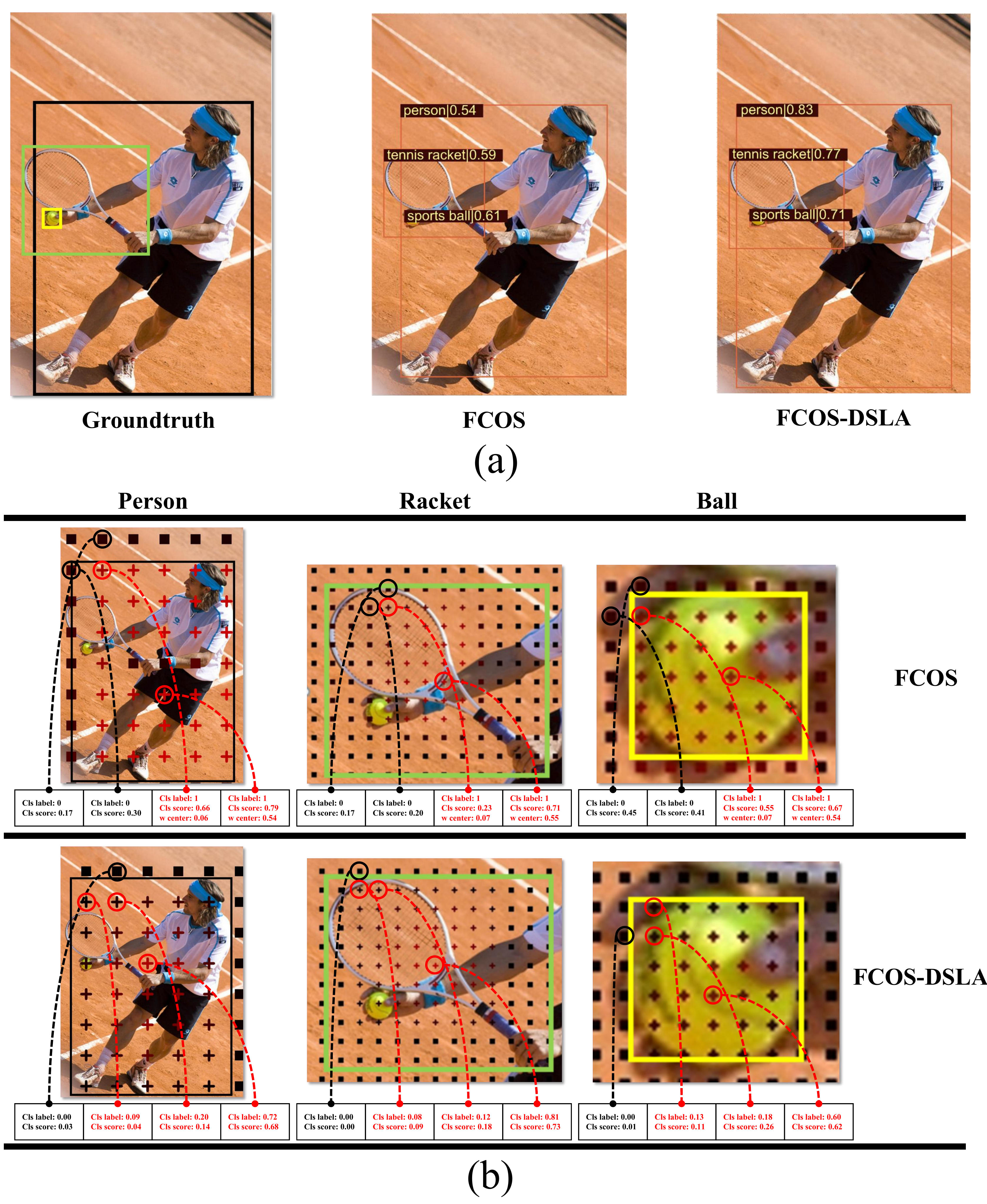}
\caption{Visualization of positive and negative division and detection results. In (a), the objects on the image are denoted by \textit{gt} bounding boxes, which are predicted on different feature maps due to their different scales. Middle and right images are the detection results of FCOS and our method. In (b), the images on the top row illustrate the positive and negative division of FCOS while the images on the bottom row illustrate the division of our method. In the images, square and cross respectively stand for assigned positive and negative samples, with the colour indicates predicted classification confidence belonging to positive samples. A redder colour indicates a higher confidence and a blacker colour a lower confidence. The scores of several adjacent samples are annotated, and the samples with the highest confidence scores are shown as well.} 
\label{fig:AG_example}
\end{figure}

FCOS makes a prediction for each location on multi-level feature maps. If the center of the receptive field (RF) \cite{RF:Dumoulin} of the location falls into a \textit{gt} box, the distances from the center to the four sides of the box are computed. If the maximum distance lies within the predefined range, then the location is set as a positive sample and is required to regress the box. An example of the sample division is illustrated in Figure \ref{fig:AG_example}(b). The input image is fed into a trained FCOS model to obtain the classification score whose value is indicated by the colour. We note that it is often the case that adjacent locations are assigned completely different labels. We consider the classification scores of these locations in Figure \ref{fig:AG_example}(b). On the feature maps predicting ``racket'' and ``ball'', the adjacent locations have similar scores but are assigned with different labels. Clearly, this does not match our expectations. We call this problem as \textbf{classification inconsistency}. We also observe that on the feature map predicting ``person'', the scores are discrepant. This is due to the different strides of feature maps. We suggest that, classification inconsistency is caused by adjacent samples which have similar RFs but are assigned with totally different supervisions. The inconsistency would prevent the detector from learning more effective object representations and thus degrades the performance as shown in Figure \ref{fig:AG_example}(a).
The introduction of centerness score solves the problem caused by similar classification scores to some extent. However, centerness score is only used in the ranking process of NMS during inference and inconsistency still exists in the training. 

\begin{figure}[htbp]
\centering
\includegraphics[width=2.5in]{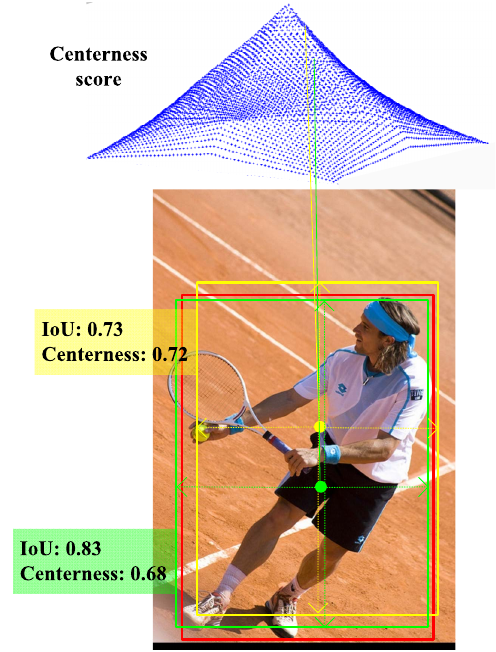}
\caption{Demonstrative case of the misalignment between centerness score and localization accuracy. The red bounding boxes denote the ground-truth box, while the yellow and green bounding boxes are both detection results yielded by locations with higher and lower centerness scores, respectively. The misalignment may lead to accurately localized bounding boxes (in green color) being suppressed by less accurate one (in yellow color) in the NMS procedure.} 
\label{fig:centerness_example}
\end{figure}

In FCOS, centerness is used to estimate localization quality. The score is predicted and combined with classification confidence as final ranking score of NMS.
Despite the improvements, centerness score would not be completely appropriate for the estimation of localization quality. In Fig.~\ref{fig:centerness_example}, it is found that the bounding box predicted by the location with higher centerness score contrarily has smaller overlap with the \textit{gt}. This is mainly because the location with higher centerness score (yellow point) is on the background and thus it cannot capture sufficient semantic information to predict an accurate bounding box. For objects with varied appearances, fixed centerness score could not always provide the reliable estimation of localization quality. We call the issue \textbf{quality estimation inconsistency}. The introduction of centerness score may lead to unexpected small ground-truth label, which makes a set of \textit{gt} boxes hard to be recalled. Researchers \cite{IoUnet:Jiang, GFL:Li, IoU-aware:Wu} suggest that IoU score would be superior than centerness score. However, IoU score is constantly changing in the entire training process and is extremely low in the early stage of training. Such dynamic values would make the training process vibrate substantially.

To improve the performance of anchor-free detectors, this paper proposes the Dynamic Smoothed Label Assignment (DSLA) method. In DSLA, the concept of centerness originally developed in FCOS is utilized but with two improvements, i.e., core zone and interval relaxation. Core zone is defined for each \textit{gt} box to maintain an sufficient confidence score, by which the problem of neglecting a true object due to its small confidence score is resolved. Interval relaxation is introduced to overcome drastic changes of assigned labels. On the basis, the labels are smoothed to continuous values in $[0, 1]$, with which a steady transition between positive and negative samples is accomplished. IoU score is dynamically calculated in the training process and is coupled with centerness score to provide a reasonable estimation of localization quality. Consequently, dynamic smooth labels are deduced to supervise the classification branch. Figure \ref{fig:AG_example}(b) illustrates the division results of FCOS equipped with DSLA. Compared with those of FCOS, the predicted classification scores are more consistent with the assigned targets. The detection results are compared in Figure\ref{fig:AG_example}(a). It can be observed that, the bounding boxes predicted by DSLA are more precise thanks to the settlement of inconsistency. And the confidence scores of the true objects increase evidently. 
With DSLA, the classification branch predicts not only the category label but also the localization quality, which could be directly used as ranking scores of NMS. The quality branch that commonly used in anchor-free detection is not needed any more. Consequently, the architecture of the detector becomes more concise and the consistency of training and inference is maintained.

The contributions of the paper are summarized as follows:
\begin{itemize}
    \item The inconsistencies of classification and quality estimation are pointed out and analyzed. Dynamic smooth label assignment is proposed to address the problems.
    \item 
    Interval relaxation strategy is proposed and combined with the improved centerness score. The assigned label is smoothed to a continuous value in $[0, 1]$, with which a steady transition between positive and negative samples is accomplished. 
    \item
    IoU score is dynamically calculated and coupled with the smooth label to supervise the classification branch of the detector. Under the supervision of DSLA, the inconsistency problems are greatly alleviated.
    \item The proposed method is applied to popular anchor-free detectors. Comprehensive experiments are carried out on MS COCO \cite{COCO:Lin} to demonstrate the effectiveness.
\end{itemize}

The remainder of this paper is organized as follows.
Section 2 briefly reviews prior work, and Section 3 describes
our approach. Experimental results are provided in Section 4. And the work is concluded in Section 5.

\section{Related Works}
\subsection{Anchor-free detectors}
The concept of anchor is initially proposed by Ren \textit{et al.} \cite{FasterRCNN:Ren} and is commonly utilized in one-stage \cite{SSD:Liu, Focalloss:Lin} and two-stage detectors \cite{FasterRCNN:Ren, MaskRCNN:He}. For anchor-based detectors, some hyper-parameters are introduced to describe the number and shapes of anchor boxes. Moreover, the parameters are specific to different detection tasks, hindering further application of anchor-based detectors. YOLOv1 \cite{YOLO:Redmon} is a popular anchor-free detector, which directly predicts bounding boxes at the points near the center of objects. However, YOLOv1 suffers from low recall rate and the following work YOLOv2 \cite{YOLO9000:Redmon} returns to anchor-based fashion.

RepPoints \cite{RepPoints:Yang} represents object as a set of sample points and utilizes deformable convolution \cite{DCN:Dai} to learn appearance features of objects. In \cite{FSAF:Zhu}, the anchor-free module FSAF is proposed and applied to RetinaNet. FSAF predicts the best feature level to train each instance. A number of anchor-free detectors solve object detection in a per-pixel prediction fashion. FCOS \cite{FCOS:Tian}, CenterNet \cite{CenterNet:Zhou} and FoveaBox \cite{FoveaBox:Kong} formulate object detection in a per-pixel prediction fashion. In these works, Feature Pyramid Network (FPN) \cite{FPN:Lin} is utilized to fuse multi-level feature maps on which the distances to the four sides of \textit{gt} boxes are predicted. In pedestrian detection, CSP \cite{CLP:Liu} includes two branches to respectively predict the center and the scale. Bounding boxes are automatically generated with the predictions and an uniform aspect ratio.

Another family of anchor-free detectors adopt keypoint-based pipeline.  CornerNet \cite{CornerNet:Law} detects a pair of corners of a bounding box and groups them to form the final detected bonding box. Inheriting from CornerNet, CenterNet \cite{CenterNet:Duan} adds an extra branch to predict the center keypoint to identify the correctness of each bounding box. ExtremeNet \cite{ExtremeNet:Zhou} detects four extreme points and one center point of objects. The five keypoints are grouped into a bounding box if they are geometrically aligned. The above detectors adopt the Hourglass network as the backbone, which is computationally expensive. Besides, they require much more complicated post-processing to group the keypoints belonging to the same instance. Anchor-free detectors have led the recent trend of object detection due to their simplicity and high performance. To improve the performance of anchor-free detectors, an efficient dynamic smooth assignment is proposed in the paper.

\subsection{Label assignment strategy in object detection}
Recent advances in object detection have shown great improvements with innovative architecture designs \cite{FPN:Lin, Aug-FPN:Guo, GFR:Guo}, normalization methods \cite{GN:Wu, DIN:Jing}, training objectives \cite{Focalloss:Lin, B_loss:Shuang, DIoU:Zheng}, additional supervisions \cite{C_supervision:Peng} and more contextual information \cite{A_block:Chen}. How to define positive and negative samples is an important issue in the training of object detectors, which greatly affects learning efficiency. As pointed by Zhang \textit{et al.} \cite{ATSS:Zhang}, the essential difference between anchor-based and anchor-free detection is actually how to define positive and negative training samples. FreeAnchor \cite{FreeAnchor:Zhang} constructs top-k anchor candidates for each \textit{gt} box and then learns to perform positive/negative division based on the detection-customized likelihood. ATSS \cite{ATSS:Zhang} selects a set of closest anchors for each \textit{gt} and divides positive and negatives anchors based on the statistics of IoU values. In \cite{GFL:Li}, IoU score is integrated into the classification branch to maintain the consistency of training and inference. PAA \cite{PAA:Kim} uses Gaussian Mixture Model (GMM) to fit the joint distribution of anchor scores to estimate the possibility of each anchor being a positive or a negative sample. OTA \cite{OTA:Ge} formulates the assigning procedure as an optimal transport problem and finds the globally best assignment solution by solving the problem at minimal transportation costs. Autoassign \cite{Autoassign:Zhu} automatically determines positive/negative samples in both spatial and scale dimensions. Notably, most of the approaches, such as FreeAnchor, ATSS, PAA and OTA, are anchor-based, which encounter the trouble of anchor setting. PAA and OTA also involve quite a few numerical iterations to find the optimal assignment solution in each training stage, increasing computational cost. OTA and Autoassign do not take smoothed labels into consideration, hindering to obtain better assignment solutions. 
The inconsistency problems commonly encountered by anchor-free detectors have not been sufficently explored. This paper proposes an efficient label assignment strategy which explores the usefulness of dynamic smooth labels.

\section{Our method}
\subsection{Centerness based smooth label assignment}
FCOS uses five levels of feature maps $\{ P_3, P_4, P_5, P_6, P_7\}$ to detect objects with different scales. For feature level $i$, if a feature point falls into a \textit{gt} box and the maximum distance lies within the predefined range, then it is defined as positive and is required to regress the box. The positive/negative sample division rule in FCOS can be summarized as: 
\begin{equation}
\label{equation_orig_head_assign}
    {\rm head^i} = \left\{
    \begin{array}{l} 1.0 \quad
    \begin{array}{ll}  m_{i-1} < \max (l^*, t^*, r^*, b^*) \le m_{i}, and \\ \min (l^*, t^*, r^*, b^*) > 0, \end{array} \\
    0.0 \quad otherwise
    \end{array}
    \right.
\end{equation}
where ${\rm head^i}$ is the assigned score of a certain point on the $i-$th level feature map, $i \in \{1,2,3,4,5\}$. $\{m_j\}_{j=0}^5$ are the hyper-parameters defining the ranges, which are set as $\{0, 64, 128, 256, 512, \infty\}$, respectively. $l^*$, $r^*$, $t^*$ and $b^*$ are the distances from the location to the four sides of the bounding box.
We denote $\max (l^*, t^*, r^*, b^*)$ as $\max$ for short in the following. 

The classification confidence predicted by the model is combined with centerness score as the final ranking score of NMS.
Centerness score \footnote{Centerness mentioned in the paper is referred to the one proposed in FCOS, unless explicitly specified.} is calculated as follows:
\begin{equation}
\label{centerness_calculation}
\begin{aligned}
        {\rm centerness} = \sqrt{\frac{\min (l^*, r^*)}{\max (l^*, r^*)} \times \frac{\min (t^*, b^*)}{\max (t^*, b^*)}}, 
\end{aligned}
\end{equation}
As the name suggests, 
centerness measures the location how close to the box center. If the location and the box center completely overlap, the location is assigned the highest score 1.0. Then the score gradually decays to 0.0 as the location deviates from the center. The introduction of centerness score is to suppress low-quality predicted bounding boxes produced by locations far away from the center of an object. Centerness is consistent with the concept of Effective Receptive Field (ERF) \cite{ERF:Luo}, based on which the researchers point out that feature points would pay more attention to the center areas of the RFs. The points that locate far away from the center of the box can not capture sufficient semantic information to represent the object. In this sense, the spacial distribution of centerness scores are rational. The utilization of centerness greatly boosts the performance of FCOS. As discussed above, the division rule would cause inconsistency. Then, a question arises naturally\textbf{---}whether centerness can be treated as the classification confidence supervision to deal with the inconsistency?

\begin{figure*}[t!]
\centering
\includegraphics[width=3.0in]{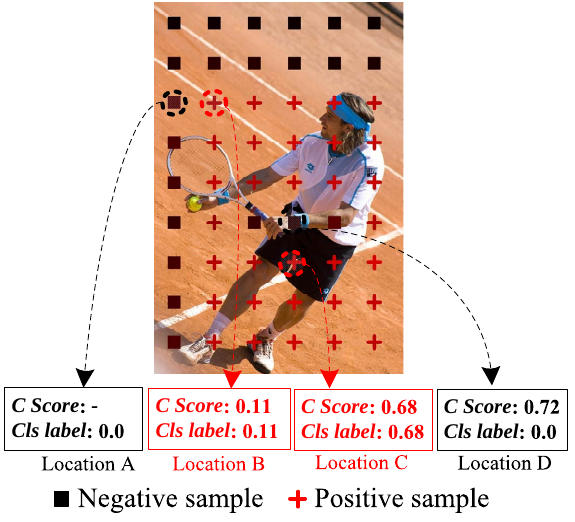}
\caption{Demonstration of label smoothing with centerness. Two groups of adjacent locations are taken as examples. Their centerness scores (abbreviated with \textit{C Score} in the figure) and the assigned labels smoothed by centerness scores (denoted as \textit{Cls Label} in the figure) are annotated. The centerness score of $A$ is skipped since it is not in the \textit{gt} box. Location $B$ and $C$ are positive samples while $A$ and $D$ are negative.}
\label{fig:Smooth_Center}
\end{figure*}

We report in Table \ref{tabel:ablation_study} that only merging centerness branch to classification branch (denoted as Impr 0) brings the improvement of $0.3 \%$ mAP. This supports our analysis.
However, when smoothing the label with centerness, the following problems need to be considered. As Fig.~\ref{fig:Smooth_Center} shows, measured by the centerness, the positive sample $B$ is assigned a relatively low target, thereby its transition to the negative sample $A$ is smooth. Nevertheless, the positive sample $C$ is assigned a much higher target. But for its adjacent sample $D$, the target is sharply dropped to 0.0, potentially leading to inconsistency again. Besides, centerness scores are hardly assigned with the largest value 1.0 due to the small possibilities that the locations exactly hit the box centers. This would lead to unexpected small confidence scores, which makes a possible set of \textit{gt} boxes extremely hard to be recalled.

To deal with the problems, the interval relaxation strategy is firstly proposed.
For $m_j, j=1,...,4$, new lower and upper bounds are determined as:  
\begin{equation}
\label{eqution_new_bound}
\begin{aligned}
& m_j^l =& m_j \times (1-\kappa),\\
& m_j^u =& m_j \times (1+\kappa),
\end{aligned}
\end{equation}
where $m_j^l$ and $m_j^u$ are lower and upper bounds related to $m_j$, $\kappa \in [0, 1)$ is the adjustment factor. Note that we set $m_0^l = m_0$ and $m_5^u = m_5$ to avoid meaningless bounds. Based on the bounds, interval relaxation is conducted, in which the head scores are formulated as:
\begin{equation}
\label{equation_new_head_assign}
    {\rm head_s^i} = \left\{
    \begin{aligned}
    & 1.0, & m_{i-1} < \max \le m_{i}\\
    & \frac{m_{i-1}-\max}{m_{i-1}-m_{i-1}^l}, & m_{i-1}^l < \max \le m_{i-1}\\
    & \frac{\max - m_{i}}{m_{i}^u-m_{i}}, & m_{i} < \max \le m_{i}^u\\
    & 0.0, & otherwise
    \end{aligned}
    \right.
\end{equation}
According to Eq.~\ref{eqution_new_bound} and Eq.~\ref{equation_new_head_assign}, the head scores lie between the new and old bounds are smoothed linearly, and $\kappa$ is a hyper-parameter which will be further investigated in experiments. 

Then, for each \textit{gt} box, the core zone is defined. Given a \textit{gt} box denoted as $(b_l, b_t, b_r, b_b)$, where $(b_l, b_t)$ and $(b_r, b_b)$ are the coordinates of the left-top and right-bottom corners, respectively. For the feature map with stride $s$, the area $(z_l, z_t, z_r, z_b)$ is defined as:
\begin{equation}
\begin{aligned}
\label{equation_core_zone}
    z_l &= \max(\ 0.5(b_l+b_r) - s/2,\ b_l\ ), \\
    z_r &= \min(\ 0.5(b_l+b_r) + s/2,\ b_r\ ), \\
    z_t &= \max(\ 0.5(b_t+b_b) - s/2,\ b_t\ ), \\
    z_b &= \min(\ 0.5(b_t+b_b) + s/2,\ b_b\ ),
\end{aligned}
\end{equation}

The area is called \textit{core zone}, which is a square with the side length of the stride. FCOS allows to predict the same \textit{gt} box on different levels of feature maps. A \textit{gt} box may have multiple core zones with different sizes due to the different strides of feature maps. For locations falling into core zones, the centerness scores are directly set to 1.0 instead of being calculated by  Eq.~\ref{centerness_calculation}. Then, the centerness is reformulated as:
\begin{equation}
\label{equation_modified_centerness}
    {\rm centerness_s} = \left\{
    \begin{aligned}
    & \sqrt{\frac{\min (l^*, r^*)}{\max (l^*, r^*)} \times \frac{\min (t^*, b^*)}{\max (t^*, b^*)}},& C_{{\rm P}} \notin Z\\
    & 1.0, & C_{{\rm P}} \in Z
    \end{aligned}
    \right.
\end{equation}
where $C_{{\rm P}}$ denotes the location and $Z$ is the core zone.
Feature points are regularly aligned with the interval of stride in both $x-$ and $y-$ directions. It is worth noting that, for each core zone, there is at least one point falling into it, to which the highest score 1.0 is assigned. Compared with Eq.~\ref{centerness_calculation}, higher confidence score could be achieved with the improved centerness.

With the computed head scores,
the label for each location across the heads is smoothed as
\begin{equation}
\label{equation_final_score}
{\rm label_s} = {\rm centerness_s} \times {\rm head_s}.
\end{equation}
Compared with directly using centerness to smooth the label, the proposed method is more efficient to make a smooth transition between positive and negative samples.
 
Note that a location may be assigned to more than one \textit{gt} boxes. FCOS simply chooses the box with minimal area as its target, which is hand-crafted and sub-optimal. The smooth label provides a natural way to solve the ambiguity problem. The box which is assigned with the highest score would be chosen as the prediction target.


\subsection{IoU based dynamic label assignemnt}

The regression branch and the classification branch are trained independently. However, in inference, the output scores from the classification branch are used as the confidence to rank the boxes predicted by the regression branch. This leads to the misalignment between training and inference. Besides, it is not completely appropriate for the classification branch to be fully supervised by the labels smoothed with centerness score. Fixed centerness score could not adapt to various shapes and attitudes of objects to always provide a reasonable estimation of localization quality.

In previous publications \cite{IoUnet:Jiang, IoU-aware:Wu}, an individual branch is added for predicting IoU scores that are used for NMS in inference. In our work, IoU scores are dynamically calculated by comparing the predicted bbox and \textit{gt} bboxes during training and are coupled into the classification branch.

IoU score is coupled with centerness score by multiplication to supervise the classification branch. Centerness score serves as the prior to stabilize the training, especially in early stage while the dynamically updated IoU score can rationally reshape the score distribution for better NMS ranking. Besides, the strategy could make the two branches interactive to maintain the consistency of training and inference. And the network architecture is more concise compared with FCOS by removing the centerness branch.

Concretely, IoU-score coupling is activated only for positive points. In each training iteration, the IoU scores between predicted boxes and \textit{gt} boxes are computed online, then are multiplied by centerness scores to obtain the final scores. The final score is deduced by:
\begin{equation}
\label{equation_final_score_reform}
{\rm label_d} = {\rm label_s} \times {\rm IoU_s},
\end{equation}
where ${\rm IoU_s}$ denotes the IoU score. The cause of classification inconsistency and how DSLA settles the inconsistencies are analyzed in Appendix.

\subsection{Application of DSLA to anchor-free detectors}

\begin{figure*}[t!]
\centering
\includegraphics[width=3.4in]{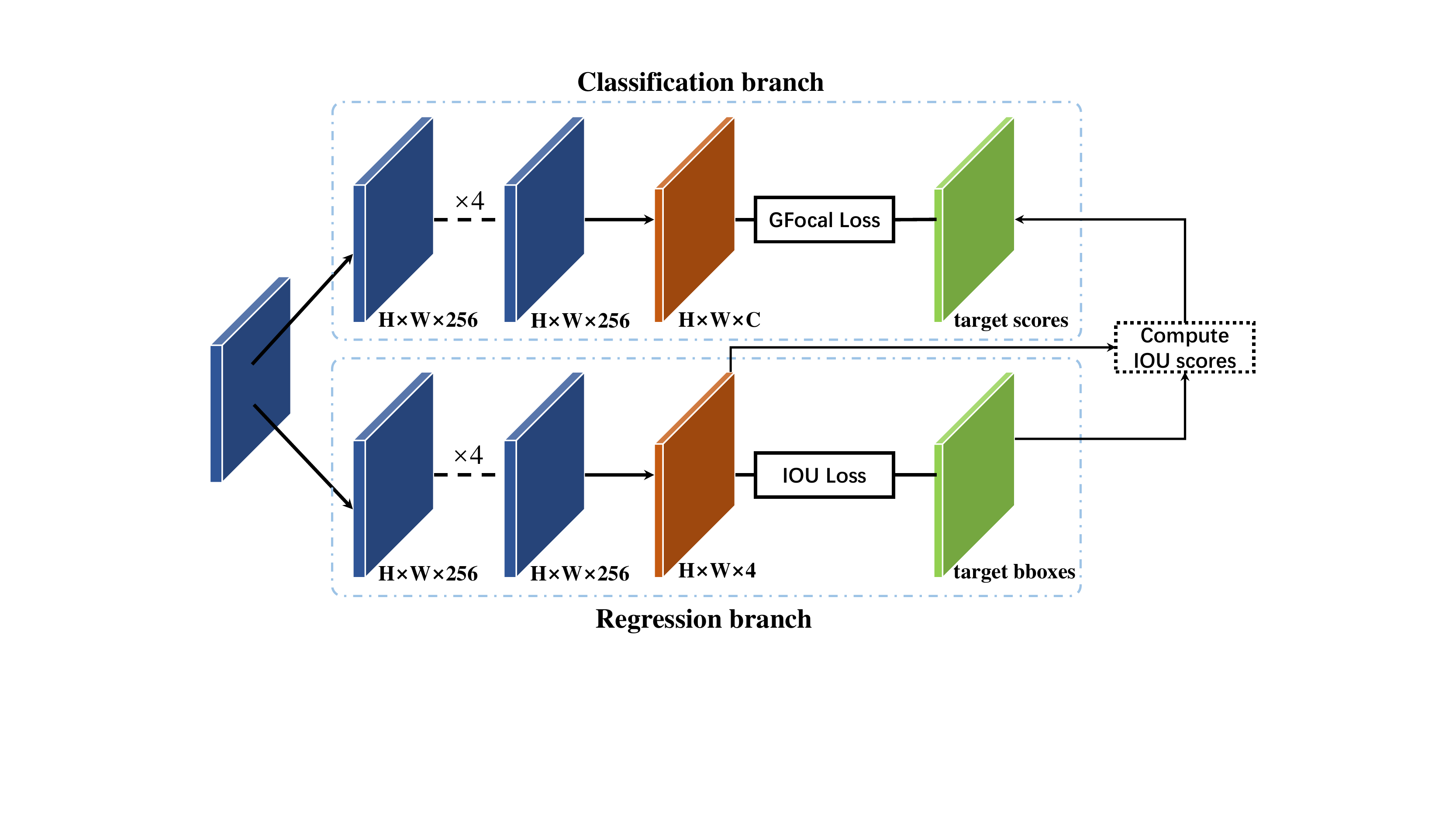}
\caption{Head structure of our method. Only two branches, classification branch and regression branch, are reserved. The inner structures are the same as ones in FCOS. $H$ and $W$ are height and width of feature maps, $C$ is the number of classes. IoU scores are produced by computing the IoU between the predicted boxes and the \textit{gt} boxes.}
\label{fig:SLL-head}
\end{figure*}

We adopt the similar network with FCOS except that the centerness branch is omitted, as shown in Fig.~\ref{fig:SLL-head}. Our method only includes the classification branch and the regression branch. As detailed in previous sections, the smoothed labels in which both of improved centerness score and IoU score are involved are used to supervise the classification branch. Similar to FCOS, in the regression branch, the distances to the four sides of the box are regressed. The loss function is
\vspace{-1mm}
\begin{eqnarray}
    L(\{p_{x,y}\}, \{t_{x,y}\})=\frac{\lambda_1}{N_{pos}} \sum_{(x,y)} L_{cls} (p_{x,y}, c^{*}_{x,y}) \nonumber\\
    +\frac{\lambda_2}{N_{pos}}\sum_{x,y}\mathrm{1}\{c^{*}_{x,y}>0\}L_{reg}(t_{x,y}, t^{*}_{x,y})
\vspace{-3mm}
\end{eqnarray}
where, $L_{cls}$ is generalized focal loss as in \cite{GFL:Li} and $L_{reg}$ is the IoU loss. $N_{pos}$ denotes the number of positive samples and $\lambda_1$ and $\lambda_2$ are the hyper-parameters to balance the weights between $L_{reg}$ and $L_{cls}$. $\mathrm{1}\{c^{*}_i>0\}$ is the indicator function being 1 if $C^{*}_i>0$ and 0 otherwise.

\section{Experiments}

Our experiments are conducted on the large-scale detection benchmark COCO \cite{COCO:Lin}. The dataset contains 115k, 5k and 20k images for train, validation and test, respectively. We train detectors on train set and the results on the validation set are reported. Mean Average Precision (MAP) is adopted as the evaluation metric. We use the evaluation code provided by COCO official.

\subsection{Implementation Details}

In the experiments, we use ResNet50 with FPN as the default backbone and neck. As in FCOS, five levels of feature maps $\{ P_3, P_4, P_5, P_6, P_7\}$ are used to detect different scales of objects. $P_3$, $P_4$ and $P_5$ are produced by the backbone feature maps $C_3$, $C_4$ and $C_5$. $P_6$ and $P_7$ are produced by $P_5$ and $P_6$, respectively. The same hyper-parameters with FCOS are used. In the training procedure, we utilize stochastic gradient descent (SGD) for optimization. The learning rate is set to 0.01 at the beginning with the batch size of 16, and the learning rate is degraded by multiplying 0.1 at epoch 8 and 11 under $1\times$ scheduler. Weight decay is set as 0.0001 and the momentum is 0.9. The backbone is pre-trained on ImageNet and the weights of the layers in neck and head are initialized as in FCOS. Following traditional routines, input image is resized to make their shorter side being 800 pixels while keeping their longer side being less or equal to 1333 pixels. Input image is randomly flipped horizontally with the probablity of $0.5$. In inference, we directly use output scores from classification branch for NMS, which is different from FCOS. The post-processing is exactly the same with FCOS. Our method is implemented using PyTorch and mmdetection \cite{MMDet:Chen}.

\subsection{Ablation study and analysis}

To demonstrate the effectiveness, sample division results derived by FCOS and the proposed method are visualized and compared in Figure \ref{fig:AG_example}. Several adjacent samples and the samples with the highest confidence scores are annotated in the figure. We note the cases in Figure \ref{fig:AG_example}(b) that adjacent locations are assigned completely different labels, for which FCOS predicts similar scores. The problem which is referred to as classification inconsistency is greatly alleviated by DSLA. With DSLA, the predicted scores are more consistent with the assigned targets, leading to more accurate bounding boxes as shown in Figure \ref{fig:AG_example}(a).

Ablation experiments are conducted where the original version of FCOS\cite{FCOS:Tian} is used as the baseline. For a fair comparison, the improvements are not utilized, such as DCN \cite{DCN:Dai}, GIoU loss \cite{GIoU:Zhang}, multi-scale train and test. 
The detectors in the comparison are trained under $1\times$ scheduler (i.e., 12 epochs) and the proposed ablation variations have the same settings as the baseline besides the improved parts. In our method, there exists only one hyper-parameter $\kappa$. We conduct the following experiments by setting $\kappa=0.2$. Subsequently, the effect of $\kappa$ is investigated independently.


\begin{table}[h!]
\scriptsize
    \centering
    \caption{Performances of detectors with different configurations. Core-Zone, Inter-Relax and IoU-Couple denote the improvements of centerness score, interval relaxation and IoU score coupling, respectively. The method FCOS in the first row denotes the FCOS baseline which has three branches. Impr 0 in the second row represents the detector in which the centerness branch is directly merged into the classification branch and the GFL loss is used to supervise the branch. Both of FCOS and Impr 0 do not make any use of the improvements.}
    \begin{tabular}{ccccc}
        \toprule
        Methods & \textbf{Core-Zone} & \textbf{Inter-Relax} & \textbf{IoU-Couple} & AP  \\
        \midrule
        FCOS & \XSolidBold & \XSolidBold & \XSolidBold & 0.366 \\
        Impr 0 & \XSolidBold & \XSolidBold & \XSolidBold & 0.369 \\
        Impr 1 & \CheckmarkBold  & \XSolidBold & \XSolidBold & 0.372 \\
        Impr 2 & \XSolidBold & \CheckmarkBold  & \XSolidBold & 0.371  \\
        Impr 3 & \XSolidBold & \XSolidBold & \CheckmarkBold & 0.373 \\
        Impr 4 & \CheckmarkBold  & \CheckmarkBold  & \XSolidBold & 0.374 \\
        Impr 5 & \CheckmarkBold  & \XSolidBold & \CheckmarkBold & 0.379 \\
        Impr 6 & \XSolidBold & \CheckmarkBold  & \CheckmarkBold & 0.376 \\
        \textbf{DSLA} & \CheckmarkBold  & \CheckmarkBold  & \CheckmarkBold & \textbf{0.381} \\
        \bottomrule
    \end{tabular}
    \vspace{-1mm}
    \vspace{-3mm}
    \label{tabel:ablation_study}
\end{table}

\textbf{Two branches vs. three branches:}
The centerness branch is directly merged into the classification branch and the Generalized Focal Loss (GFL) \cite{GFL:Li} is used to supervise the branch (Impr 0 in Table \ref{tabel:ablation_study}). Without other changes, the improvement outperforms the baseline by $0.3\%$ mAP ($36.6\%$ \textit{v.s.} $36.9\%$). This indicates that merging centerness branch to classification branch not only simplifies network architecture but also improves the performance.

\textbf{Three improvements:} 
To analysis the importance of each component, we gradually add core zone, interval relaxation and IoU score coupling on the ResNet-50 FPN FCOS baseline. Different detectors are deduced by using different configurations. The experimental results are reported in Table \ref{tabel:ablation_study}. The introduction of centerness core zone (Impr 1) boosts the performance from $36.9\%$ mAP to $37.2\%$ mAP. Interval relaxation changes the distribution of centerness score for the heads. Using interval relaxation brings the improvement of $0.2\%$ mAP (Impr 4). This proves the superiority of the proposed centerness representation. With all of the three improvements, our method achieves the highest $38.1\%$ mAP, outperforming the baseline by $1.5 \%$ mAP. It can be concluded from Table  \ref{tabel:ablation_study} that, every single improvement can stably boost the performance. Meanwhile, IoU score coupling is marginally more efficient than the other two improvements.

In Impr 0, replacing centerness score by IoU score deduces the mAP from $36.6\%$ to $35.2\%$. This is because, IoU score constantly changes in the entire training process and is extremely low in the early stage. Learning such values is inefficient. As the prior, centerness score would force the network to converge to a better local optima. The experiments demonstrate the effectiveness of DSLA. Meanwhile, the proposed coupling strategy would be helpful to further boost the performance of other anchor-free detectors.

\textbf{Effect of hyper-parameter $\kappa$:}
DSLA introduces the hyper-parameter $\kappa$. Obviously, $\kappa$ affects the transition between positive and negative samples and the label assignment across different levels of heads. We conduct several experiments to study the effect of $\kappa$. Different values of $\kappa$ in $[0.1, 0.2, 0.3, 0.4]$ are used to train the detector and the results are shown in Table \ref{tabel:hyper-effect}. From the results, we observe that, the performance is insensitive to $\kappa$. Setting $\kappa=0.2$ achieves the highest performance and thus it is adopted in all the experiments.

\begin{table}[h!]
\footnotesize
    \centering
    \caption{Effect of the hyper-parameter $\kappa$. }
    \begin{tabular}{cccc}
        \toprule
        $\kappa$ & $AP$ & $AP_{50}$ & $AP_{75}$  \\
        \midrule
        0.1 & 0.379 & 0.568 & 0.410\\
        \midrule
        0.2  & 0.381 & 0.568 & 0.413 \\
        \midrule
        0.3  & 0.378  & 0.566 & 0.410\\
        \midrule
        0.4  & 0.377  & 0.564 & 0.407  \\
        \bottomrule
    \end{tabular}
    \vspace{-2mm}
    \vspace{-3mm}
    \label{tabel:hyper-effect}
\end{table}

\subsection{Application of DSLA to popular anchor-free detectors}

Popular anchor-free detectors are chosen as the baselines. The representative anchor-free detector, FCOS, is firstly chosen. Its improved version \cite{FCOSv2:Tian} further boosts the performance, which is denoted as FCOSv2 in the comparison experiments. The classical anchor-free detector, FoveaBox \cite{FoveaBox:Kong}, is chosen as well. For a fair comparison, we follow all the training settings of the original experiments in the baselines \cite{FCOS:Tian, FCOSv2:Tian, FoveaBox:Kong} except for the improved components. ResNet-50 and ResNet-101 backbones are adopted and the comparison results are provided in Table \ref{tabel:map_coco_res}. We observe from the table that, the improvements by using DSLA commonly exceed $1\%$ mAP. Especially, with the ResNet-101, DSLA improves FCOSv2 from $41.5\%$ mAP to $44.1\%$ mAP. The improvements are obvious. 

\begin{figure*}[htbp]
\centering
\includegraphics[width=4.0in]{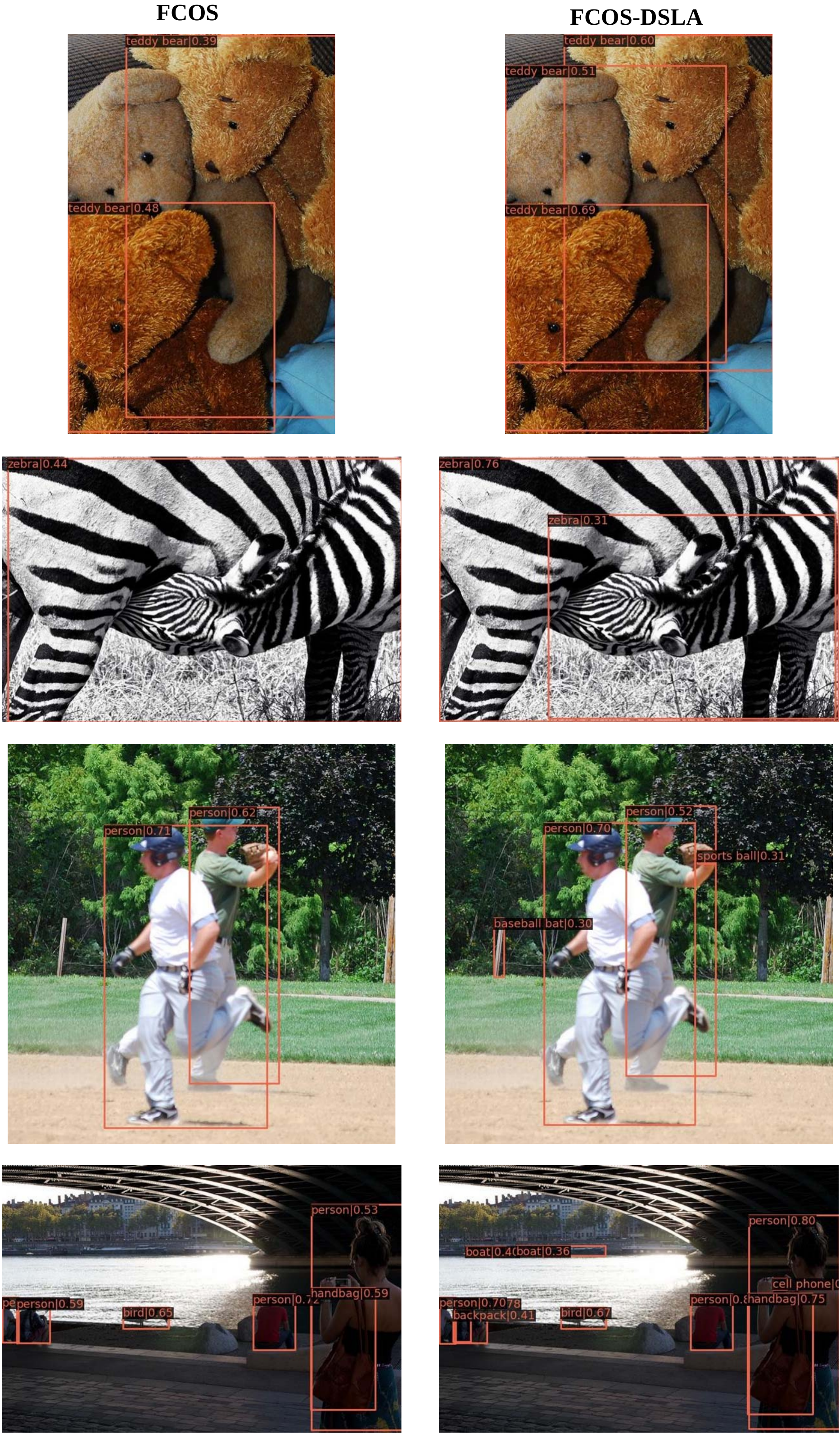}
\caption{FCOS \textit{vs} FCOS w / DSLA. Both models adopt ResNet-101 backbone and are trained with COCO trainval35k dataset. The left column contains the results from the conventional FCOS and the right
column is from FCOS equipped with DSLA. Bounding boxes with score of 0.3 or higher are drawn.}
\label{fig:FCOS_comp}
\end{figure*}

\begin{figure*}[htbp]
\centering
\includegraphics[width=4.0in]{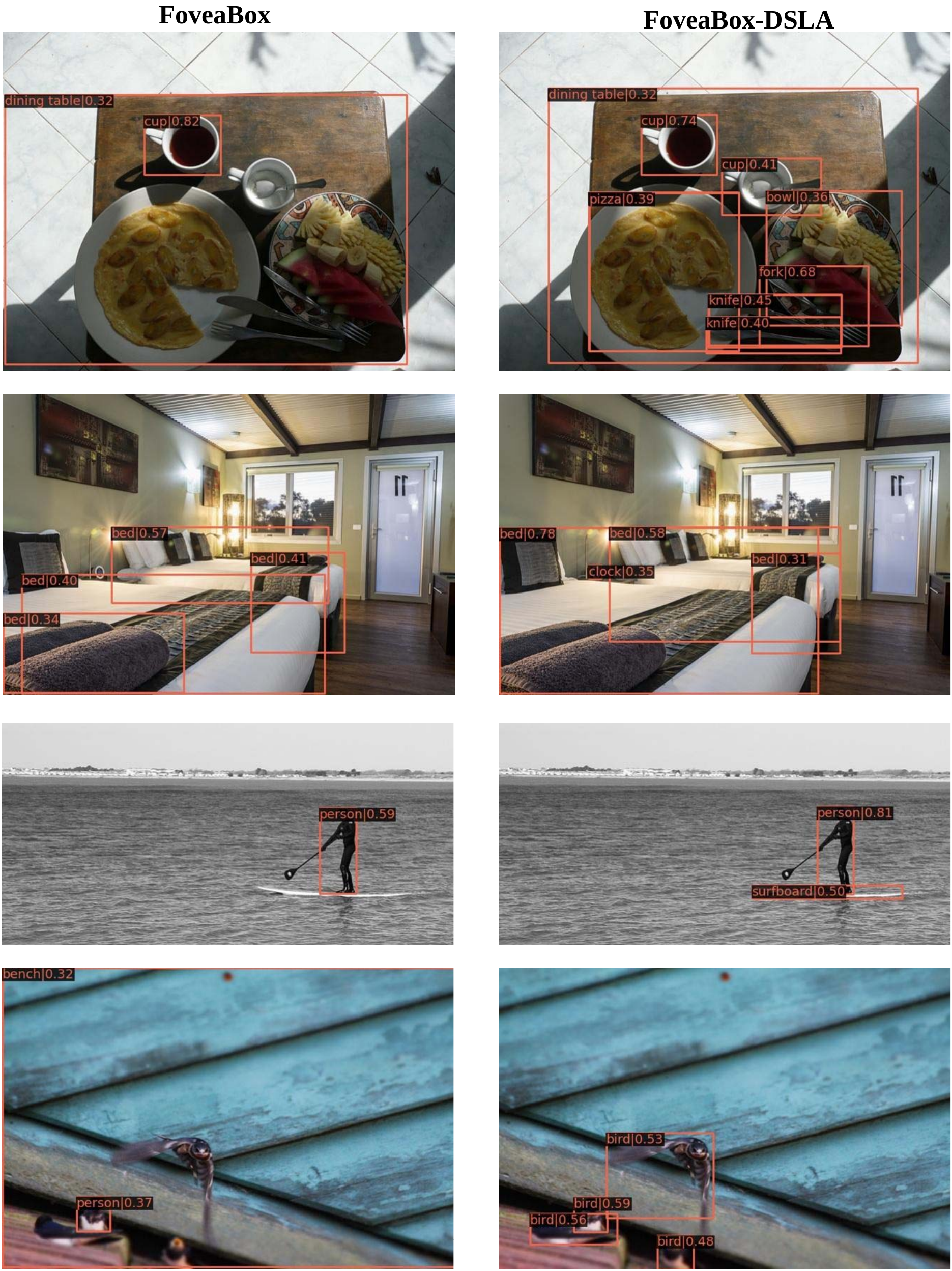}
\caption{FoveaBox \textit{vs} FoveaBox w / DSLA. Both models adopt ResNet-101 backbone and are trained with COCO trainval35k dataset. The left column contains the results from the conventional FoveaBox and the right column is from FoveaBox equipped with DSLA. Bounding boxes with score of 0.3 or higher are drawn.}
\label{fig:FoveaBox_comp}
\end{figure*}

Several examples of detection results are visualized in Figure \ref{fig:FCOS_comp} and Figure \ref{fig:FoveaBox_comp} for better illustration. As shown in Rows 1 and 2 in the figures, DSLA could predict more precise bounding boxes. Meanwhile, as shown in Rows 3 and 4 in the figures, baseline models cannot detect several small objects. Nevertheless, after applying DSLA, these objects are successfully detected. This is because DSLA increases the confidence score with the core zone strategy and thus improves the recall rate of small objects. From Table \ref{tabel:map_coco_res}, we can also observe that DSLA performs better on small objects compared with the baselines.
The experiments demonstrate the effectiveness of DSLA. 



\begin{table}[h!]
\footnotesize
    \centering
    \caption{Detection results on MS COCO. }
\begin{tabular}{l|l|ccc|ccc}
    \toprule
    Method & Backbone and neck & AP & $AP_{50}$ & $AP_{75}$ & $AP_S$ & $AP_M$ & $AP_L$ \\
    \hline
    FCOS\cite{FCOS:Tian} & ResNet-50-FPN & 0.366 & 0.560 & 0.389 & 0.209 & 0.403 & 0.472 \\
    FCOS\cite{FCOS:Tian} & ResNet-101-FPN & 0.391 & 0.583 & 0.421 & 0.227 & 0.433 & 0.503 \\
    \textbf{FCOS w / DSLA} & \textbf{ResNet-50-FPN} & \textbf{0.381} & \textbf{0.568} & \textbf{0.413} & \textbf{0.212} & \textbf{0.422} & \textbf{0.506} \\
    \textbf{FCOS w / DSLA} & \textbf{ResNet-101-FPN} & \textbf{0.399} & \textbf{0.588} & \textbf{0.434} & \textbf{0.227} & \textbf{0.438} & \textbf{0.531} \\
    \hline
    FCOSv2\cite{FCOSv2:Tian} & ResNet-50-FPN & 0.386 & 0.574 & 0.414 & 0.223 & 0.425 & 0.498 \\
    FCOSv2\cite{FCOSv2:Tian} & ResNet-101-FPN & 0.415 & 0.607 & 0.450 & 0.244 & 0.448 & 0.516 \\
    \textbf{FCOSv2 w / DSLA} & \textbf{ResNet-50-FPN} & \textbf{0.404} & \textbf{0.603} & \textbf{0.442} & \textbf{0.227} & \textbf{0.448} & \textbf{0.529} \\
    \textbf{FCOSv2 w / DSLA} & \textbf{ResNet-101-FPN} & \textbf{0.441} & \textbf{0.623} & \textbf{0.482} & \textbf{0.263} & \textbf{0.480} & \textbf{0.577} \\
    \hline
    FoveaBox\cite{FoveaBox:Kong} & ResNet-50-FPN & 0.365 & 0.560 & 0.386 & 0.205 & 0.399 & 0.477 \\
    FoveaBox\cite{FoveaBox:Kong} & ResNet-101-FPN & 0.386 & 0.579 & 0.411 & 0.216 & 0.425 & 0.504 \\
    \textbf{FoveaBox w / DSLA} & \textbf{ResNet-50-FPN} & \textbf{0.375} & \textbf{0.572} & \textbf{0.404} & \textbf{0.207} & \textbf{0.412} & \textbf{0.496} \\
    \textbf{FoveaBox w / DSLA} & \textbf{ResNet-101-FPN} & \textbf{0.398} & \textbf{0.596} & \textbf{0.429} & \textbf{0.226} & \textbf{0.434} & \textbf{0.523} \\
    \bottomrule
    \end{tabular}
    \vspace{-2mm}
    \vspace{-3mm}
    \label{tabel:map_coco_res}
\end{table}

\subsection{Comparison with state-of-the-art detectors}
Comparison studies are carried out with state-of-the-art detectors. Table \ref{tabel:compare_SOTA} provides the best reported performances of the detectors with ResNet-50 backbone. Notably, CornerNet \cite{CornerNet:Law} and CenterNet \cite{CenterNet:Duan} are not involved in Table \ref{tabel:compare_SOTA} as they adopt a different type of backbone, Hourglass. We observe from the table that, DSLA achieves comparable performance to GFL ($42.7\%$ \textit{vs} $42.9\%$) and is superior than all other detectors with
the same backbone. Deformable Convolutional Networks (DCN) further boosts the performance of DSLA to $44.4\%$ mAP. Compared with FCOS, DSLA achieves a significant improvement of $2.1\%$ mAP. Additionally, higher performances are achieved when DSLA is equipped with stronger backbones. With the backbone of X-101-64x4d-DCN, $48.1\%$ mAP is achieved. Using Swin-S as the backbone and following the default setting to adopt 
$3 \times$ scheduler, the performance of $49.2\%$ mAP is achieved. The proposed DSLA greatly improves the performances of anchor-free detectors.

\begin{table}[h!]
\footnotesize
    \centering
    \caption{Performance comparison with state-of-the-art detectors on MS COCO. \textbf{R}: ResNet. \textbf{X}: ResNeXt. \textbf{DCN}: Deformable Convolutional Network.}
    \begin{tabular}{l|l|c|ccc|ccc}
    \toprule
    Method & Backbone & Epoch & AP & $AP_{50}$ & $AP_{75}$ & $AP_S$ & $AP_M$ & $AP_L$ \\
    \hline
    \textit{multi-stage:} & & & & & \\
    Cascade R-CNN\cite{CascadeRCNN:Cai}  & R-50 & 20 & 0.419 & 0.600 & 0.459 & 0.232 & 0.449 & 0.559 \\
    Grid R-CNN\cite{GridRCNN:Cai} & R-50 & 25 & 0.404 & 0.585 & 0.436 & 0.227 & 0.439 & 0.530 \\
    Libra R-CNN\cite{LibraRCNN:Pang} & R-50 & 12 & 0.383 & 0.595 & 0.419 & 0.221 & 0.420 & 0.485 \\
    TridentNet\cite{TridentNet:Li} & R-50 & 36 & 0.402 & 0.598 & 0.435 & 0.217 & 0.444 & 0.562 \\
    RepPoints\cite{RepPoints:Yang} & R-50 & 24 & 0.386 & 0.596 & 0.416 & 0.225 & 0.422 & 0.504 \\
    \hline
    \textit{one-stage anchor-based:} & & & & & \\
    RetinaNet \cite{Focalloss:Lin} & R-50 & 24 & 0.374 & 0.567 & 0.396 & 0.200 & 0.407 & 0.497 \\
    ATSS \cite{ATSS:Zhang} & R-50 & 12 & 0.394 & 0.576 & 0.428 & 0.236 & 0.429 & 0.503 \\
    PAA \cite{PAA:Kim} & R-50 & 24 & 0.416 & 0.598 & 0.453 & 0.244 & 0.450 & 0.546 \\
    GFL \cite{GFL:Li} & R-50 & 24 & 0.429 & 0.612 & 0.465 & 0.273 & 0.469 & 0.533 \\
    \hline
    \textit{one-stage anchor-free:} & & & & & \\
    Autoassign \cite{Autoassign:Zhu} & R-50 & 12 & 0.404 & 0.596 & 0.437 & 0.227 & 0.441 & 0.529 \\
    FoveaBox \cite{FoveaBox:Kong} & R-50 & 24 & 0.404 & 0.612 & 0.429 & 0.245 & 0.444 & 0.521 \\
    FCOSv2 \cite{FCOSv2:Tian} & R-50 & 24 & 0.385 & 0.577 & 0.410 & 0.219 & 0.428 & 0.486\\
    FCOSv2 \cite{FCOSv2:Tian}& R-50-DCN & 12 & 0.423 & 0.611 & 0.454 & 0.244 & 0.459 & 0.558\\
    FSAF \cite{FSAF:Zhu} & R-50 & 12 & 0.374 & 0.568 & 0.398 & 0.204 & 0.411 & 0.488\\
    \midrule
    \textit{Our method:} & & & & & \\
    DSLA & R-50 & 24 & 0.427 & 0.603 & 0.462 & 0.269 & 0.468 & 0.549\\
    DSLA & R-50-DCN & 24 & 0.444 & 0.625 & 0.484 & 0.266 & 0.477 & 0.593\\
    DSLA & X-101-64x4d-DCN & 24 & 0.481 & 0.668 & 0.522 & 0.299 & 0.518 & 0.629\\
    DSLA & Swin-S & 36 & 0.492 & 0.681 & 0.535 & 0.326 & 0.525 & 0.641 \\ 
    \bottomrule
    \end{tabular}
    \vspace{-1mm}
    \vspace{-3mm}
    \label{tabel:compare_SOTA}
\end{table}

\section{Conclusion}
The inconsistency problems suffered by prevalent anchor-free detectors are investigated in the paper. A novel method, DSLA, is proposed to continuously boost the performance of anchor-free detectors by addressing the inconsistency problems. Interval relaxation strategy is proposed and is combined with the improved representation of centerness to make the transition between positive and negative samples smoother. Dynamic IoU score is coupled with the classification branch to provide a reasonable estimation of localization quality. As a result, dynamic smooth labels are deduced, with which inconsistencies is greatly alleviated. DSLA naturally integrates the centerness branch suggested in FCOS into the classification branch to make the architecture simpler and to maintain the consistency of training and inference. Extensive experiments conducted on MS COCO validate the effectiveness of DSLA. 
For further improvement of DSLA, more optimal combination of centerness score and IoU score deserves in-depth investigation. In the future, extending the smooth label assignment strategy to anchor-based detector would be a promising research topic. Additionally, inconsistency is a general problem that would be encountered in other tasks such as semantic segmentation and category classification. Using smooth label to alleviate the inconsistencies would be meaningful.

\section*{Acknowledgements}
This work is supported in part by the National Key Research and
Development Program of China under Grant 2018YFB1306303, and in part by the National Natural Science Foundation of China under Grant 61773374, 61702323 and 62172268, and in part by the Major Basic Research Projects of Natural Science Foundation of Shandong Province under Grant ZR2019ZD07, and in part by the Guangdong Basic and Applied Basic Research Foundation under Grant 2020B1515120050.

\begin{appendix}
\section{Inconsistency analysis}
\begin{figure*}[t!]
\centering
\includegraphics[width=4.5in]{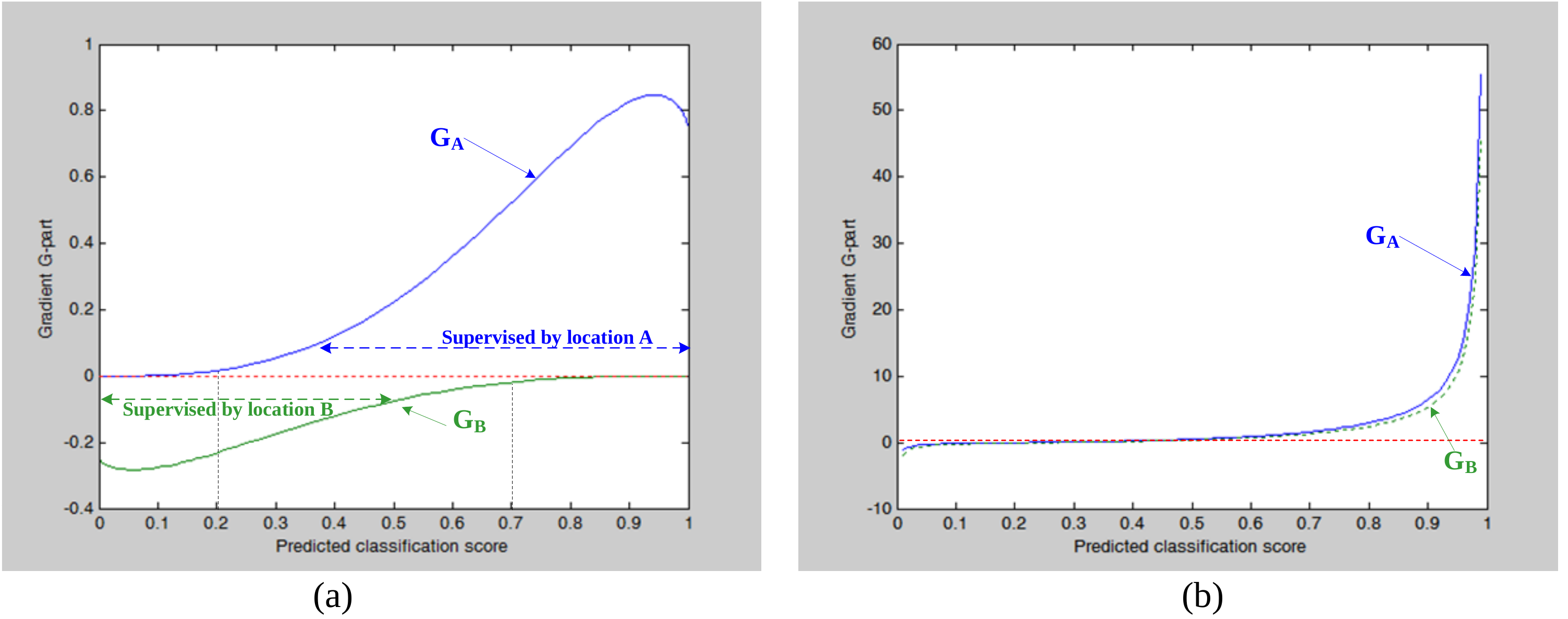}
\caption{G-part curves. In (a), discrete labels are assigned and the loss is calculated by FL \cite{Focalloss:Lin}. In (b), smooth labels are assigned and the loss is calculated by GFL \cite{GFL:Li} }
\label{fig:Quan_Grad}
\end{figure*}

The convolutional network $N_e$ is represented by a list of composed convolutional layers: $N_e(X)=F_M \odot F_{M-1} \odot \cdots \odot F_1(X)=\bigodot_{j=1 \cdots M}F_j(X)$, where $F_i$ denotes the $i-th$ convolutional layer, $X$ is the input tensor. Suppose that two adjacent locations $A$ and $B$ are identified as positive and negative samples, respectively. Their RFs denoted as $RF_A$ and $RF_B$ are similar and the centers of their RFs are $(x_A, y_A)$ and $(x_B, y_B)$. The classification loss of anchor-free detector is defined as
\begin{equation}
CL=\frac{1}{N_{pos}} \sum_{(x,y)}FL_{cls}(p_{\theta}(x,y), c^{*}(x,y))
\end{equation}
where, $FL_{cls}$ is focal loss as in \cite{Focalloss:Lin}, $N_{pos}$ denotes the number of positive samples, $p_{\theta}(x,y)$ is the classification score predicted by the network with parameter $\theta$ for location $(x, y)$ and $c^{*}(x,y)$ is the target. Then, the following relationships are obtained
\begin{eqnarray}
p_{\theta}(x_A,y_A) & = & S(N_e(RF_A)) \\
p_{\theta}(x_B,y_B) & = & S(N_e(RF_B))
\end{eqnarray}
where $S(\cdot)$ is the $sigmoid$ function defined as $S(x)=\frac{1}{1+e^{-x}}$. 
In the following, $p_{\theta}(x_A,y_A)$ and $p_{\theta}(x_B,y_B)$ are simplified as $p^A_{\theta}$ and $p^B_{\theta}$, respectively. The gradients of the two locations are derived by using the chain rule of compound function 
\begin{eqnarray}
\left. \nabla_{\theta} CL \right|_{(x_A, y_A)}  & = &  \bigg[ \alpha \gamma (1-p^A_{\theta}) ^{\gamma - 1} \log (p^A_{\theta}) \nonumber \\
& & \underbrace {- \alpha (1-p^A_{\theta}) ^{\gamma} \frac{1}{p^A_{\theta}} \bigg] S(x)(1-S(x))}_{G_A} \underbrace{\left. \frac{\partial N_e} {\partial \theta}\right|_{(x_A, y_A)}}_{D_A} \\
\left. \nabla_{\theta} CL \right|_{(x_B, y_B)} & = &  \bigg[ -(1-\alpha) \gamma {p^B_{\theta}} ^{\gamma - 1} \log (1-p^B_{\theta}) \nonumber \\
& & \underbrace { + (1-\alpha) {p^B_{\theta}} ^{\gamma} \frac{1}{1-p^B_{\theta}} \bigg] S(x)(1-S(x))}_{G_B} \underbrace{\left. \frac{\partial N_e}{\partial \theta}\right|_{(x_B, y_B)}}_{D_B}
\end{eqnarray}
where $\alpha$ and $\gamma$ are hyper-parameters and are set to $0.25$ and $2$ as in \cite{Focalloss:Lin}. For each location, the derived gradient is divided into two parts, G-part and D-part. For example, $\left. \nabla_{\theta} CL \right|_{(x_A, y_A)}$ can be expressed as $G_A \cdot D_A$, where $G_A$ and $D_A$ respectively represent the G-part and D-part of the gradient. 
$G_A$ and $G_B$ play important role in gradient back-propagation. Figure \ref{fig:Quan_Grad}(a) shows how $G_A$ and $G_B$ change as the functions of predicted scores. We note from the figure that, $G_A$ and $G_B$ have different signs. If the predicted score is greater than 0.7 (or smaller than 0.2), the gradient derived by $A$ (or $B$) is high which dominates the learning. If the score lies in $[0.2, 0.7]$, the network is trained with both of the two opposing gradients. If the score is close to $0.5$, the total gradient vanishes by counteraction. The classification inconsistency might be attributed to \textbf{inconsistent gradients (IG)} when the network tries to balance the totally different supervisions but with similar inputs. Obviously, the existing of IG would lead to learning inefficiency and degrades the performance. 

Using smooth label as the supervision, the classification is optimized by GFL \cite{GFL:Li}. The gradients are uniformly derived as
\begin{eqnarray}
\left. \nabla_{\theta} CL \right|_{(x, y)}  & = &  2(y-p_{\theta})\left[(1-y)\log (1-p_{\theta})+y \log (p_{\theta})\right] \nonumber \\ 
& & \underbrace{+ (y-p_{\theta})^2 \left [(1-y) \frac{1}{1-p_{\theta}} - y \frac{1}{p_{\theta}}\right]}_{G} \underbrace{\left. \frac{\partial N_e} {\partial \theta}\right|_{(x, y)}}_{D}
\end{eqnarray}
where, $y$ is the assigned smooth label and $p_{\theta}$ is the simplification of $p_{\theta}(x, y)$. The curves of $G_A$ and $G_B$ are shown in Figure \ref{fig:Quan_Grad}(b). Smooth change of the assigned labels only causes small deviation of the gradients. By comparing the curves in Figure \ref{fig:Quan_Grad}(a) and Figure \ref{fig:Quan_Grad}(b), we found that gradient consistency is guaranteed. As a result, the predicted classification scores are more consistent with training objectives, as illustrated in Figure \ref{fig:AG_example}. Thus, smooth label greatly alleviates classification inconsistency by addressing IG.

Figure \ref{fig:centerness_example} illustrates the quality estimation inconsistency. The accurately localized green box is suppressed by the less accurate yellow box in NMS. By coupling with the IoU score, green box has a higher confidence than yellow box and is maintained after NMS. Compared with only centerness score, coupling with IoU score would provide more reasonable estimation of localization quality. Therefore, the deduced dynamic smooth label overcomes the limitation of quality estimation inconsistency.

\end{appendix}
\bibliography{DSLA_PR}

\begin{thebibliography}{10}
\expandafter\ifx\csname url\endcsname\relax
  \def\url#1{\texttt{#1}}\fi
\expandafter\ifx\csname urlprefix\endcsname\relax\def\urlprefix{URL }\fi
\expandafter\ifx\csname href\endcsname\relax
  \def\href#1#2{#2} \def\path#1{#1}\fi

\bibitem{ResNet:He}
K.~He, X.~Zhang, S.~Ren, J.~Sun, Deep residual learning for image recognition,
  in: Proceedings of the IEEE Conference on Computer Vision and Pattern
  Recognition, 2016, pp. 770--778.

\bibitem{RCNN:Girshick}
R.~Girshick, J.~Donahue, T.~Darrell, J.~Malik, Rich feature hierarchies for
  accurate object detection and semantic segmentation, in: Proceedings of the
  IEEE Conference on Computer Vision and Pattern Recognition, 2014, pp.
  580--587.

\bibitem{Point2Seq:Xue}
Y.~Xue, J.~Mao, M.~Niu, H.~Xu, M.~Bi~Mi, W.~Zhang, X.~Wang, X.~Wang, Point2seq:
  Detecting 3d objects as sequences, arXiv preprint arXiv:2203.13394.

\bibitem{FCN:Long}
E.~Shelhamer, J.~Long, T.~Darrell, Fully convolutional networks for semantic
  segmentation, IEEE Transactions on Pattern Analysis and Machine Intelligence
  39 (2017) 640--650.

\bibitem{FGCN:Yang}
Y.~Yang, Z.~Feng, M.~Song, X.~Wang, Factorizable graph convolutional networks,
  in: Advances in Neural Information Processing Systems, 2020, pp.
  20286--20296.

\bibitem{DGCN:Yang}
Y.~Y. Yang, J.~Qiu, M.~Song, D.~Tao, X.~Wang, Distilling knowledge from graph
  convolutional networks, in: Proceedings of the IEEE/CVF Conference on
  Computer Vision and Pattern Recognition, 2020, pp. 7072--7081.

\bibitem{Cross-modality:Huang}
Z.~Huang, Z.~Zeng, H.~Yupan, B.~Liu, D.~Fu, J.~Fu, Seeing out of the box:
  end-to-end pre-training for vision-language representation learning, in:
  Proceedings of the IEEE/CVF Conference on Computer Vision and Pattern
  Recognition, 2021, pp. 12971--12980.

\bibitem{QFasterRCNN:Zhang}
J.~Zhang, Z.~Zhang, H.~Su, W.~Zou, X.~Gong, F.~Zhang, Quality inspection based
  on quadrangular object detection for deep aperture component, IEEE
  Transactions on Systems, Man, and Cybernetics: Systems 51 (2021) 1938--1948.

\bibitem{PSLA:Guo}
C.~Guo, B.~Fan, J.~Gu, Q.~Zhang, S.~Xiang, V.~Prinet, C.~Pan, Progressive
  sparse local attention for video object detection, in: Proceedings of the
  IEEE/CVF International Conference on Computer Vision, 2019, pp. 3908--3917.

\bibitem{GASTD:Wang}
F.~Wang, L.~Zhao, X.~Li, X.~Wang, D.~Tao, Geometry-aware scene text detection
  with instance transformation network, in: Proceedings of the IEEE/CVF
  Conference on Computer Vision and Pattern Recognition, 2018, pp. 1381--1389.

\bibitem{RIM:Wang}
G.~Wang, X.~Wang, F.~Fan, Bin, C.~Pan, Feature extraction by rotation-invariant
  matrix representation for object detection in aerial image, IEEE Geoscience
  and Remote Sensing Letters 14 (2017) 851--855.

\bibitem{FasterRCNN:Ren}
S.~Ren, K.~He, R.~Girshick, J.~Sun, Faster r-cnn: Towards real-time object
  detection with region proposal networks, IEEE Transactions on Pattern
  Analysis and Machine Intelligence 39 (2017) 1137--1149.

\bibitem{SSD:Liu}
W.~Liu, D.~Anguelov, D.~Erhan, C.~Szegedy, S.~Reed, C.-Y. Fu, A.~C. Berg, Ssd:
  Single shot multibox detector, in: Proceedings of European Conference on
  Computer Vision, Springer, 2016, pp. 21--37.

\bibitem{Focalloss:Lin}
T.-Y. Lin, P.~Goyal, R.~Girshick, K.~He, P.~Doll{\'a}r, Focal loss for dense
  object detection, in: Proceedings of the IEEE International Conference on
  Computer Vision, 2017, pp. 2980--2988.

\bibitem{YOLO9000:Redmon}
J.~Redmon, A.~Farhadi, Yolo9000: better, faster, stronger, in: Proceedings of
  the IEEE Conference on Computer Vision and Pattern Recognition, 2017, pp.
  7263--7271.

\bibitem{YOLO:Redmon}
J.~Redmon, S.~Divvala, R.~Girshick, A.~Farhadi, You only look once: unified,
  real-time object detection, in: Proceeding of the IEEE Conference on Computer
  Vision and Pattern Recognition, 2016, pp. 779--788.

\bibitem{CornerNet:Law}
H.~Law, J.~Deng, Cornernet: Detecting objects as paired keypoints, in:
  Proceedings of the European Conference on Computer Vision, 2018, pp.
  734--750.

\bibitem{CenterNet:Duan}
K.~Duan, S.~Bai, L.~Xie, H.~Qi, Q.~Huang, Q.~Tian, Centernet: keypoint triplets
  for object detection, in: Proceedings of the IEEE/CVF International
  Conference on Computer Vision, 2019, pp. 6568--6577.

\bibitem{FCOS:Tian}
Z.~Tian, C.~Shen, H.~Chen, T.~He, Fcos: Fully convolutional one-stage object
  detection, in: Proceedings of the IEEE/CVF International Conference on
  Computer Vision, 2019, pp. 9627--9636.

\bibitem{CenterNet:Zhou}
X.~Zhou, D.~Wang, P.~Krahenbuhl, Object as points, arXiv preprint
  arXiv:1904.07850v1.

\bibitem{FoveaBox:Kong}
T.~Kong, F.~Sun, H.~Liu, Y.~Jiang, L.~Li, J.~Shi, Foveabox: beyond anchor-based
  object detection, IEEE Transactions on Image Processing 29 (2020) 7389 --
  7398.

\bibitem{ATSS:Zhang}
S.~Zhang, C.~Chi, Y.~Yao, Z.~Lei, S.~Z. Li, Bridging the gap between
  anchor-based and anchor-free detection via adaptive training sample
  selection, in: Proceedings of the IEEE/CVF Conference on Computer Vision and
  Pattern Recognition, 2020, pp. 9759--9768.

\bibitem{OTA:Ge}
Z.~Ge, S.~Liu, Z.~Li, O.~Yoshie, J.~Sun, Ota: Optimal transport assignment for
  object detection, in: Proceedings of the IEEE/CVF Conference on Computer
  Vision and Pattern Recognition, 2021, pp. 303--312.

\bibitem{RF:Dumoulin}
V.~Dumoulin, F.~Visin, A guide to convolution arithmetic for deep learning,
  arXiv preprint arXiv:1603.07285.

\bibitem{IoUnet:Jiang}
B.~Jiang, R.~Luo, J.~Mao, T.~Xiao, Y.~Jiang, Acquisition of localization
  confidence for accurate object detection, in: Proceedings of the European
  Conference on Computer Vision, 2018, pp. 784--799.

\bibitem{GFL:Li}
X.~Li, W.~Wang, L.~Wu, S.~Chen, X.~Hu, J.~Li, J.~Tang, J.~Yang, Generalized
  focal loss: Learning qualified and distributed bounding boxes for dense
  object detection, in: Advances in Neural Information Processing Systems,
  2020, pp. 21002--21012.

\bibitem{IoU-aware:Wu}
S.~Wu, X.~Li, X.~Wang, Iou-aware single-stage object detector for accurate
  localization, Image and Vision Computing 97 (2020) 103911.

\bibitem{COCO:Lin}
T.-Y. Lin, M.~Maire, S.~Belongie, J.~Hays, P.~Perona, D.~Ramanan,
  P.~Doll{\'a}r, C.~L. Zitnick, Microsoft coco: Common objects in context, in:
  Proceedings of European Conference on Computer Vision, Springer, 2014, pp.
  740--755.

\bibitem{MaskRCNN:He}
K.~He, G.~Gkioxari, P.~Dollar, R.~Girshick, Mask r-cnn, in: Proceedings of the
  IEEE International Conference on Computer Vision, 2017, pp. 2980--2988.

\bibitem{RepPoints:Yang}
Z.~Yang, S.~Liu, H.~Hu, L.~Wang, S.~Lin, Reppoints: Point set representation
  for object detection, in: Proceedings of the IEEE/CVF International
  Conference on Computer Vision, 2019, pp. 9656--9665.

\bibitem{DCN:Dai}
J.~Dai, H.~Qi, Y.~Xiong, Y.~Li, G.~Zhang, H.~Hu, W.~Yichen, Deformable
  convolutional networks, in: Proceedings of the IEEE/CVF International
  Conference on Computer Vision, 2017, pp. 764--773.

\bibitem{FSAF:Zhu}
C.~Zhu, Y.~He, M.~Savvides, Feature selective anchor-free module for
  single-shot object detection, in: Proceedings of the IEEE/CVF Conference on
  Computer Vision and Pattern Recognition, 2019, pp. 840--849.

\bibitem{FPN:Lin}
T.-Y. Lin, P.~Doll{\'a}r, R.~Girshick, K.~He, B.~Hariharan, S.~Belongie,
  Feature pyramid networks for object detection, in: Proceedings of the IEEE
  International Conference on Computer Vision, 2017, pp. 936--944.

\bibitem{CLP:Liu}
W.~Liu, S.~Liao, W.~Ren, W.~Hu, Y.~Yu, High-level semantic feature detection: A
  new perspective for pedestrian detection, in: Proceedings of the IEEE/CVF
  Conference on Computer Vision and Pattern Recognition, 2019, pp. 5182--5191.

\bibitem{ExtremeNet:Zhou}
Z.~Xingyi, Z.~Jiacheng, K.~Philipp, Bottom-up object detection by grouping
  extreme and center points, in: Proceedings of the IEEE/CVF Conference on
  Computer Vision and Pattern Recognition, 2019, pp. 850--859.

\bibitem{Aug-FPN:Guo}
C.~G. Guo, B.~Fan, Q.~Zhang, S.~Xiang, C.~Pan, Augfpn: improving multi-scale
  feature learning for object detection, in: Proceedings of the IEEE/CVF
  Conference on Computer Vision and Pattern Recognition, 2020, pp.
  12592--12601.

\bibitem{GFR:Guo}
Z.~Shen, H.~Shi, J.~Yu, H.~Phan, R.~Feris, L.~Cao, D.~Liu, X.~Wang, T.~Huang,
  M.~Savvides, Improving object detection from scratch via gated feature reuse,
  in: Proceedings of the British Machine Vision Conference, 2019.

\bibitem{GN:Wu}
Y.~Wu, K.~He, Group normalization, in: Proceedings of the European Conference
  on Computer Vision, 2018, pp. 3--19.

\bibitem{DIN:Jing}
Y.~Jing, X.~Liu, Y.~Ding, X.~Wang, E.~Ding, M.~Song, S.~Wen, Dynamic instance
  normalization for arbitrary style transfer, in: Proceedings of the AAAI
  Conference on Artificial Intelligence, 2020, pp. 4369--4376.

\bibitem{B_loss:Shuang}
K.~Shuang, Z.~Lyu, J.~Loo, W.~Zhang, Scale-balanced loss for object detection,
  Pattern Recognition 117 (2021) 107997.

\bibitem{DIoU:Zheng}
Z.~Zheng, P.~Wang, W.~Liu, J.~Li, R.~Ye, D.~Ren, Distance-iou loss: faster and
  better learning for bounding box regressionl feature pyramid network, in:
  Proceedings of the AAAI Conference on Artificial Intelligence, 2020, pp.
  12993--13000.

\bibitem{C_supervision:Peng}
J.~Peng, H.~Wang, S.~Yue, Z.~Zhang, Context-aware co-supervision for accurate
  object detection, Pattern Recognition 121 (2022) 108199.

\bibitem{A_block:Chen}
C.~Chen, J.~Yu, Q.~Ling, Sparse attention block: Aggregating contextual
  information for object detection, Pattern Recognition.

\bibitem{FreeAnchor:Zhang}
X.~Zhang, F.~Wan, C.~Liu, R.~Ji, Q.~Ye, Freeanchor: Learning to match anchors
  for visual object detection, in: Advances in Neural Information Processing
  Systems, 2019, pp. 147--155.

\bibitem{PAA:Kim}
K.~Kim, H.~S. Lee, Probabilistic anchor assignment with iou prediction for
  object detection, in: Proceedings of the European Conference on Computer
  Vision, 2020, pp. 355--371.

\bibitem{Autoassign:Zhu}
B.~Zhu, J.~Wang, Z.~Jiang, f.~Zong, S.~Liu, Z.~Li, J.~Sun, Autoassign:
  differentiable label assignment for dense object detection, arXiv preprint
  arXiv:2007.03496.

\bibitem{ERF:Luo}
W.~Luo, Y.~Li, R.~Urtasun, R.~Zemel, Understanding the effective receptive
  field in deep convolutional neural networks, in: Advances in Neural
  Information Processing Systems, 2016, pp. 4898--4096.

\bibitem{MMDet:Chen}
K.~Chen, J.~Wang, J.~Pang, Y.~Cao, X.~Yu, X.~Li, S.~S. Sun, W.~Feng, Z.~Liu,
  J.~Xu, Z.~Zhang, D.~Cheng, C.~Zhu, T.~Cheng, Q.~Zhao, B.~Li, X.~L. Lu,
  R.~Zhu, Y.~Wu, J.~Dai, J.~Wang, J.~Shi, W.~Ouyang, C.~Loy, D.~Lin,
  Mmdetection: open mmlab detection toolbox and benchmark, arXiv preprint
  arXiv:1906.07155v1.

\bibitem{GIoU:Zhang}
H.~Rezatofighi, N.~Tsoi, J.~Gwak, A.~Sadeghian, I.~Reid, S.~Savarese,
  Generalized intersection over union: a metric and a loss for bounding box
  regression, in: Proceedings of the IEEE/CVF Conference on Computer Vision and
  Pattern Recognition, 2019, pp. 658--666.

\bibitem{FCOSv2:Tian}
Z.~Tian, C.~Shen, H.~Chen, T.~He, Fcos: a simple and strong anchor-free object
  detector, IEEE Transactions on Pattern Analysis and Machine Intelligence 44
  (2021) 1922--1933.

\bibitem{CascadeRCNN:Cai}
Z.~Cai, N.~Vasconcelos, Cascade r-cnn: delving into high quality object
  detection, in: Proceedings of the IEEE/CVF Conference on Computer Vision and
  Pattern Recognition, 2018, pp. 6154--6162.

\bibitem{GridRCNN:Cai}
X.~Lu, B.~Li, Y.~Yue, Q.~Li, J.~Yan, Grid r-cnn, in: Proceedings of the
  IEEE/CVF Conference on Computer Vision and Pattern Recognition, 2019, pp.
  7355--7364.

\bibitem{LibraRCNN:Pang}
J.~Pang, K.~Chen, J.~Shi, H.~Feng, W.~Ouyang, D.~Lin, Libra r-cnn: towards
  balanced learning for object detection, in: Proceedings of the IEEE/CVF
  Conference on Computer Vision and Pattern Recognition, 2019, pp. 821--830.

\bibitem{TridentNet:Li}
Y.~Li, Y.~Chen, N.~Wang, Z.~Zhang, Scale-aware trident networks for object
  detection, in: Proceedings of the IEEE/CVF Conference on Computer Vision,
  2019, pp. 6053--6062.

\end{thebibliography}

\end{document}